\documentclass[letterpaper, 10 pt, conference]{ieeeconf}  

\IEEEoverridecommandlockouts                              
\usepackage{algorithm}
\usepackage{algpseudocode}
\usepackage{amsmath}
\usepackage{times}
\usepackage{amssymb}
\usepackage{booktabs}
\usepackage{physics}
\usepackage{graphicx}
\usepackage{url}
\usepackage[table]{xcolor}
\usepackage{float}
\usepackage{dblfloatfix}
\usepackage{subfig}
\usepackage{textcomp}
\usepackage{gensymb}
\usepackage{multicol,tabularx,capt-of}
\usepackage{multirow}
\usepackage[hidelinks]{hyperref}
\usepackage[absolute,overlay]{textpos}
\usepackage{booktabs}
\usepackage{tabularx}
\usepackage{siunitx}
\usepackage{threeparttable}
\definecolor{revisionblue}{RGB}{0,0,0}
\long\def\zj#1{{\color{revisionblue}#1}}
\newenvironment{revision}{}{}
\newcommand{\valpm}[2]{%
  \ensuremath{\scalebox{0.95}{\ensuremath{#1}}%
             \,{\scalebox{0.85}{$\pm$}}\,%
             \scalebox{0.85}{\ensuremath{#2}}}%
}

\begin{document}
\title{\LARGE \bf
\zj{Wavelet Policy: Imitation Learning in the Scale Domain with World Prior Memory}
}

\author{Changchuan Yang$^{1}$, Haoxuan Xu$^{3}$, Yuhang Dong$^{1}$, Guanzhong Tian$^{1,4}$, Haizhou Ge$^{2}$, and Hongrui Zhu$^{1}$%
\thanks{*This work is supported in part by the National Natural Science Foundation of China under Grant 62303405, in part by the Ningbo Natural Science Foundation Project under Grant 2023J400, and in part by the Ningbo Key Research and Development Plan under Grant 2023Z116. This work is also supported by Hangzhou TaiWeave Robotics Co., Ltd., through experimental assistance with the industrial sewing setup and xArm6 industrial manipulator from UFactory.}%
\thanks{$^{1}$Zhejiang University}%
\thanks{$^{2}$Tsinghua University, DISCOVER Robotics}%
\thanks{$^{3}$The Hong Kong University of Science and Technology}%
\thanks{$^{4}$Corresponding author: {\tt\small gztian@zju.edu.cn}}%
}

\maketitle
\thispagestyle{empty}
\pagestyle{empty}

\begin{abstract}
Conventional visuomotor imitation learning usually predicts future robot actions directly in the time domain. Such formulations often have limited physical scene awareness and weak memory. 
In this work, we propose Wavelet Policy, a lightweight imitation learning framework that combines \zj{World Prior Memory (WPM)} with wavelet-based multi-scale action modeling. Our key idea is to encode persistent physical scene structure from static background images into compact memory tokens, which are fused into world-prior tokens and injected into the encoder during forward propagation.
Based on this memory-conditioned representation, we further perform \zj{wavelet-domain decomposition over horizon-aligned latent action tokens} and adopt a Single-Encoder Multiple-Decoder (SE2MD) architecture to model latent components at different temporal scales. The resulting latent subbands are reconstructed through inverse wavelet transform and finally projected into executable action chunks. To facilitate efficient world prior learning, we introduce a \zj{world-prior adaptation loss}, encouraging the background encoder to retain persistent scene knowledge while remaining lightweight and stable.
Extensive experiments on four simulated and six real-world robotic manipulation tasks show that Wavelet Policy consistently outperforms strong baselines. These results demonstrate that combining scale-domain action modeling with world-prior memory provides an effective and efficient solution for embodied manipulation.
\end{abstract}


\section{INTRODUCTION}
\begin{revision}
Visuomotor imitation learning~\cite{meng2026utracker, gao2025out} aims to train a policy that maps robot observations to future actions from expert demonstrations. Recent action-chunking and diffusion policies have substantially improved long-horizon manipulation by predicting short future action sequences instead of one-step controls~\cite{chi2023diffusion, fu2024mobile, zhao2023learning}. Nevertheless, most of these policies still regress the action chunk in the time domain. This representation entangles slow transport motions, contact transitions, and high-frequency alignment corrections in a single sequence, making it difficult for a lightweight decoder to specialize across different temporal scales.

A second practical limitation is scene awareness. In many tabletop and industrial manipulation tasks, important physical cues such as support surfaces, workspace boundaries, fixed fixtures, and non-interactive regions remain persistent across a rollout. A policy that relies only on the current foreground observation must repeatedly infer such static structure from frame-wise visual features. World-model-based perception~\cite{tagliabue2024tube, nematollahi2025lumos} and memory-augmented policies~\cite{torne2026mem} address this issue by maintaining richer environmental representations, but they often require large frozen backbones, recurrent memory updates, or multi-stage training.

These observations motivate an imitation policy that is both scale-aware and memory-aware while remaining compact. Wavelet transforms provide a useful inductive bias for this purpose because they decompose a sequence into temporally localized approximation and detail components with exact reconstruction. Compared with purely learned hierarchical policies~\cite{park2024hierarchical, rana2023residual} or generic multi-scale sequence modules~\cite{liu2024multi, cui2025hierarchical}, a wavelet-domain representation explicitly separates coarse trends from local refinements before decoding, without introducing an additional planner or subgoal supervision.

As illustrated in Fig.~\ref{fig:introduction}, existing approaches can be broadly categorized into two representative paradigms. VLM-based models~\cite{intelligence2026pi, gr00tn1_2025} typically depend on frozen large-scale backbones and multi-stage alignment pipelines. End-to-end models~\cite{lee2025interact, park2024hierarchical} are simpler, but usually ask one action decoder to model variations across all temporal scales. Our goal is to preserve the efficiency of end-to-end imitation learning while adding two structured priors: latent scale-domain action modeling and a compact memory of the static world.

\begin{figure}[t]
\centering
\includegraphics[width=\columnwidth]{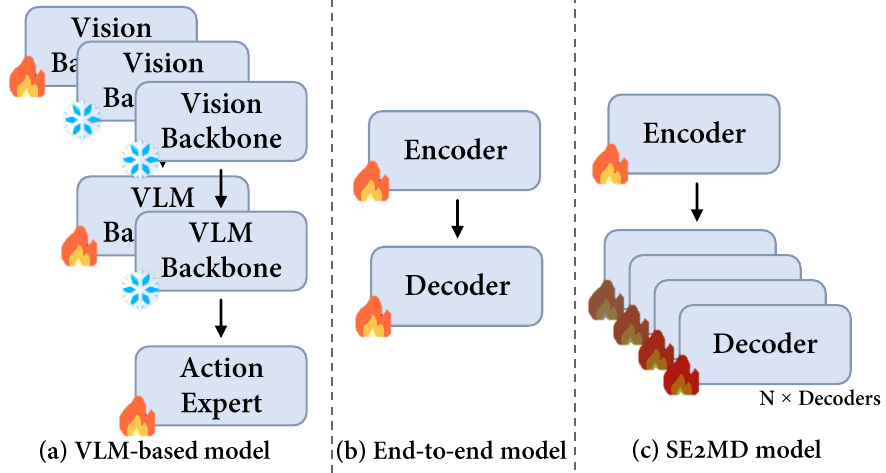}
\caption{
(a) VLM-based models rely on frozen large-scale backbones and typically require a costly multi-stage training pipeline.
(b) End-to-end models train an encoder-decoder policy to directly map observations to actions, which may limit training efficiency and representation capacity.
(c) The proposed SE2MD framework adopts a shared single encoder and multiple lightweight decoders, improving training efficiency and enabling more effective multi-scale action modeling. 
}
\vspace{-4mm}
\label{fig:introduction}
\end{figure}
To address these limitations, we propose Wavelet Policy, a lightweight framework for visuomotor imitation learning. Wavelet Policy integrates World Prior Memory (WPM) with wavelet-based~\cite{latif2024discrete, zhang2019wavelet} multi-scale action modeling. We treat action generation as a scale-structured latent prediction problem: horizon-aligned latent action tokens are decomposed into approximation and detail components, decoded by scale-specific Transformer decoders, reconstructed by an inverse wavelet transform, and finally projected to executable actions. In parallel, WPM encodes persistent static structures from background images into compact memory tokens, which are fused with online visual tokens before policy decoding. This design clarifies the data flow from observation to action while avoiding extra post-decoder filtering modules.


\begin{figure*}
\vspace{5mm}
    \centering
    \includegraphics[width=0.96\textwidth]{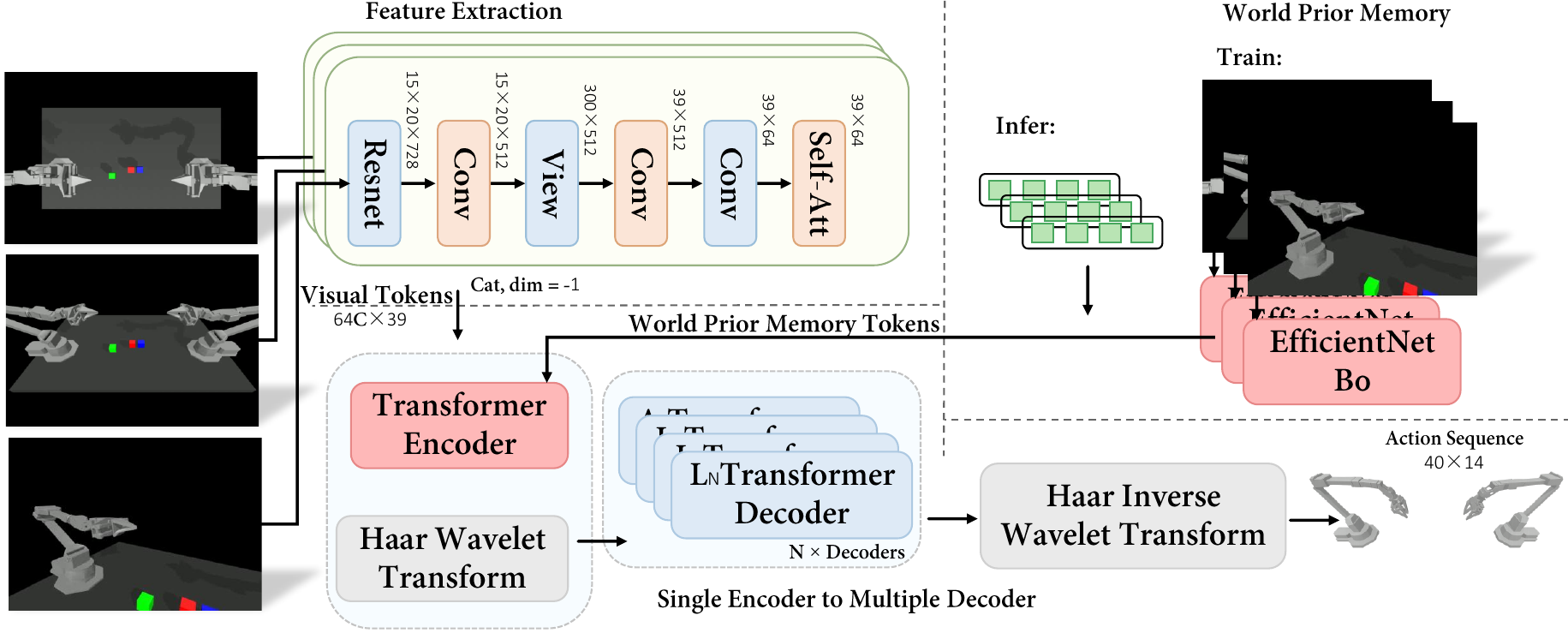}
    \caption{Wavelet Policy data flow. Multi-view online images are converted into visual tokens, while background images are encoded once into WPM tokens and cached during inference. The shared encoder processes the fused token sequence. Horizon-aligned latent action tokens are then decomposed along the horizon dimension by WT, decoded by SE2MD, reconstructed by inverse WT, and projected to the final action chunk. Here, $C$ denotes the number of cameras and $N$ denotes the number of scale decoders.}

    \label{fig:framework}
\end{figure*}

Our main contributions are summarized as follows:
\begin{itemize}
\item We introduce Wavelet Policy, a lightweight imitation learning framework that combines wavelet-based latent action modeling with WPM for improved scene awareness and efficiency.
\item We provide an end-to-end scale-domain formulation with explicit tensor shapes, Haar decomposition and reconstruction equations, and a SE2MD architecture that decodes scale-specific latent action subbands.
\item We incorporate a world-prior adaptation objective that encourages the background encoder to learn task-relevant static scene structure while keeping the cached memory compact.
\item We evaluate Wavelet Policy on simulated and real-world manipulation tasks with stronger baselines, confidence intervals, hyperparameter details, and wavelet-basis ablations.

\end{itemize}
\end{revision}

\section{Related Work}
\begin{revision}

\subsection{Visuomotor Imitation Learning}

Visuomotor imitation learning aims to learn robot policies directly from visual observations and expert demonstrations. More recent approaches improve temporal modeling by predicting action chunks or denoising action sequences with sequence models, Transformers, and diffusion-based policies~\cite{chi2023diffusion, fu2024mobile}. Representative chunk-based methods such as ACT ~\cite{zhao2023learning} and its variants ~\cite{park2024hierarchical} further improve execution by modeling future action segments rather than single-step actions. However, despite their effectiveness, most existing methods still operate primarily with coarse trajectory trends in the time domain. This makes action generation difficult, especially for manipulation tasks that require both smooth global transport and precise contact-rich adjustments.

\subsection{World-aware Perception and Memory-augmented Policies}

A growing body of work has explored incorporating explicit world knowledge into policy learning~\cite{matsuo2022deep, goff2025learning}. World-model-based perception methods improve embodied decision-making by constructing structured latent representations of the environment, while memory-augmented policies~\cite{huang2025memory, shu2025episodic} maintain persistent state information across time to enhance temporal consistency and scene awareness. These methods show that encoding static workspace geometry, object layout, and interaction context can significantly improve policy robustness in partial observations. However, stronger world awareness is often achieved at the cost of increased computational complexity, such as maintaining large latent world models, recurrent memory states, or auxiliary perception modules.

\subsection{Multi-scale and Frequency-domain Action Modeling}

Multi-scale modeling~\cite{xu2023rngdet++, zhang2023flight} has been widely studied in sequence learning because complex trajectories often contain dynamics at different temporal resolutions. In robotic manipulation, action sequences typically involve slow global motion trends together with fast local corrections, suggesting that a single-scale time-domain predictor may be insufficient. Prior work has explored hierarchical policies~\cite{park2024hierarchical}, temporal abstraction~\cite{rana2023residual}, or multi-resolution representations~\cite{cui2025hierarchical} to capture such structure. Frequency-domain and time-frequency analyses~\cite{wang2025long, sarna2024comparative} provide another natural way to separate signals into components with different temporal characteristics. Wavelet-based neural modules have also been used for image restoration, high-frequency enhancement, and trajectory forecasting~\cite{ding2024wavelet, azad2023unlocking, zou2021sdwnet, zhang2023flight}. Unlike these works, Wavelet Policy applies the wavelet transform to horizon-aligned latent action tokens in an imitation policy, then uses scale-specific decoders and exact inverse reconstruction to produce executable action chunks. This makes the scale structure explicit while avoiding additional subgoal labels or non-invertible temporal pooling.
\end{revision}


\section{Method}
\begin{revision}

\subsection{End-to-End Data Flow and Problem Formulation}

We study action-chunking imitation learning. At time step $t$, the policy receives synchronized RGB observations $\mathcal{I}_t=\{I_t^{(c)}\}_{c=1}^{N_c}$ and predicts $\hat{\mathbf{A}}_t\in\mathbb{R}^{B\times H\times D}$, where $B$, $H$, and $D$ denote batch size, horizon, and action dimension. The forward path is: online images and background images are encoded into visual tokens and WPM tokens, fused by a shared encoder, converted to horizon-aligned latent action tokens, decomposed by WT, decoded by SE2MD, reconstructed by inverse WT, and projected to actions. \zj{The wavelet transform is applied to latent future-action tokens along the horizon dimension, not to raw images or FE visual tokens.}

To make the notation explicit, superscript $(c)$ indexes camera views, superscript $(s)$ indexes wavelet decomposition levels, and subscripts $t$ and $k$ denote the current control step and the temporal index inside a horizon-aligned sequence. $N_c$ is the number of cameras; $B$, $H$, and $D$ denote batch size, action horizon, and action dimension; $L$ is the number of condensed visual tokens; $d_v$ is the per-view token width; $d$ is the shared Transformer width; and $K$ is the number of WPM tokens. We use $\mathbf{X}_t$ for online visual tokens, $\mathbf{M}_{wp}$ for WPM tokens, $\tilde{\mathbf{X}}_t$ for their concatenation, $\mathbf{H}_t$ for encoder memory, $\mathbf{Q}$ for horizon-aligned action queries, and $\mathbf{Z}_t$ for latent future-action tokens. After WT, $\mathbf{U}^{(s)}$ and $\mathbf{V}^{(s)}$ denote approximation and detail subbands, respectively; $\Pi_R$ is the projection head for subband $R$, and $g_a$ is the final action head.

\subsection{Multi-view Visual Feature Extraction}

For each view, a ResNet-style encoder produces a dense feature map. In our implementation, $480\times640$ RGB images yield a $15\times20$ grid. The tensor denoted as $15\times20\times728$ consists of a 512-channel visual map and a 216-channel auxiliary map containing positional, camera-view, and robot-state embeddings. A $1\times1$ projection maps it to $15\times20\times512$. The grid is flattened into 300 tokens, condensed to $L=39$ tokens, and projected to $d_v=64$. Multi-view features are concatenated along the embedding dimension and projected to $\mathbf{X}_t\in\mathbb{R}^{B\times L\times d}$, so the token length remains fixed as the camera number grows.

\subsection{World Prior Memory}

WPM stores static scene information from background images. For each camera, we use an object-free initial frame when available; otherwise, we use the first rollout frame. The background image is denoted by $B^{(c)}$ and encoded by EfficientNet-B0~\cite{kansal2024resnet}. View memories are fused into $K$ memory tokens and prepended to the online visual tokens:
\begin{equation}
\begin{aligned}
\mathbf{m}^{(c)} &= E_{bg}(B^{(c)}),\\
\mathbf{m}_{wp} &= W_f[\mathbf{m}^{(1)}\Vert\cdots\\
&\quad\Vert\mathbf{m}^{(N_c)}]+b_f,\\
\mathbf{M}_{wp} &= \mathrm{reshape}(W_m\mathbf{m}_{wp}+b_m),\\
\tilde{\mathbf{X}}_t &= [\mathbf{M}_{wp};\mathbf{X}_t],
\quad \mathbf{M}_{wp}\in\mathbb{R}^{B\times K\times d}.
\end{aligned}
\end{equation}
Here $E_{bg}$ denotes the lightweight background encoder, $W_f$ and $W_m$ are learned linear projections for multi-view fusion and token generation, and $b_f,b_m$ are bias terms. $\mathbf{m}^{(c)}$ is a compact per-view scene descriptor, $\mathbf{m}_{wp}$ is the fused scene vector before tokenization, and $\mathbf{M}_{wp}$ is the final memory-token sequence injected into the policy. During inference, $\mathbf{M}_{wp}$ is computed once at rollout start and then cached.

\subsection{Latent Wavelet Decomposition}

The fused tokens are processed by a shared Transformer encoder. Horizon-aligned action queries then attend to this memory:
\begin{equation}
\begin{aligned}
\mathbf{H}_t &= \mathrm{Enc}(\tilde{\mathbf{X}}_t+\mathbf{P}_{enc}),\\
\mathbf{Z}_t &= \mathrm{CrossAttn}(\mathbf{Q},\mathbf{H}_t),
\quad \mathbf{Z}_t\in\mathbb{R}^{B\times H\times d}.
\end{aligned}
\end{equation}
In this step, $\mathbf{P}_{enc}$ is the encoder positional encoding and $\mathbf{Q}$ is a learned set of horizon-aligned latent action queries. Intuitively, each row of $\mathbf{Q}$ asks the encoder memory to produce one latent token for one future step in the action horizon.

Let $\mathbf{U}^{(0)}=\mathbf{Z}_t$. Omitting batch and channel indices, a level-$s$ Haar decomposition along the horizon dimension is defined as
\begin{equation}
\begin{aligned}
\mathbf{U}^{(s)}_{k}
&=
\frac{\mathbf{U}^{(s-1)}_{2k}+\mathbf{U}^{(s-1)}_{2k+1}}{\sqrt{2}},
\\
\mathbf{V}^{(s)}_{k}
&=
\frac{\mathbf{U}^{(s-1)}_{2k}-\mathbf{U}^{(s-1)}_{2k+1}}{\sqrt{2}}.
\end{aligned}
\end{equation}
where $s=1,\ldots,S$, $k=0,\ldots,H/2^s-1$, and $\mathbf{U}^{(s)},\mathbf{V}^{(s)}\in\mathbb{R}^{B\times(H/2^s)\times d}$. $\mathbf{U}^{(s)}$ stores the coarse approximation component after $s$ levels of downsampling, while $\mathbf{V}^{(s)}$ stores the detail component added back during inverse reconstruction. With $H=40$ and $S=3$, the branch lengths are 20, 10, and 5.

\subsection{SE2MD Scale Decoding and Inverse Reconstruction}

SE2MD uses one shared encoder and one lightweight decoder for each latent wavelet subband. Let $\mathcal{R}=\{\mathbf{U}^{(S)},\mathbf{V}^{(S)},\ldots,\mathbf{V}^{(1)}\}$. For $\mathbf{R}\in\mathcal{R}$, the corresponding decoder predicts $\hat{\mathbf{R}}\in\mathbb{R}^{B\times n_R\times d}$ while attending to the same $\mathbf{H}_t$. The predicted subbands are recursively reconstructed:

\begin{equation}
\begin{aligned}
\mathbf{Y}_R &= \mathrm{Dec}_R(\mathbf{R}+\mathbf{P}_R,\mathbf{H}_t),\\
\hat{\mathbf{R}} &= \Pi_R(\mathbf{Y}_R),\\
\hat{\mathbf{U}}^{(s-1)}_{2k}
&=
\frac{\hat{\mathbf{U}}^{(s)}_{k}+\hat{\mathbf{V}}^{(s)}_{k}}{\sqrt{2}},\\
\hat{\mathbf{U}}^{(s-1)}_{2k+1}
&=
\frac{\hat{\mathbf{U}}^{(s)}_{k}-\hat{\mathbf{V}}^{(s)}_{k}}{\sqrt{2}},\\
\hat{\mathbf{A}}_t &= g_a(\hat{\mathbf{U}}^{(0)}).
\end{aligned}
\end{equation}
Here $\mathrm{Dec}_R(\cdot,\cdot)$ denotes the decoder assigned to subband $R$, $\mathbf{P}_R$ is the corresponding scale-specific positional encoding, and $\Pi_R$ maps decoder features back to the latent subband space. The final head $g_a$ converts the reconstructed latent sequence into the executable robot action chunk.

\subsection{Optimization Objective}

We train the whole model end-to-end. The primary imitation objective is an L1 action reconstruction loss on the reconstructed time-domain action chunk. \zj{Latent subbands are learned only through inverse reconstruction and the final action loss.} The WPM adaptation loss keeps the background encoder within a bounded update range:
\begin{equation}
\begin{aligned}
\mathcal{L}_{act}
&=\frac{1}{BHD}\|\hat{\mathbf{A}}_t-\mathbf{A}_t\|_1,\\
r_{bg}
&=\frac{\|\theta_{bg}-\theta_{bg}^{(0)}\|_2}
{\|\theta_{bg}^{(0)}\|_2+\epsilon},\\
\mathcal{L}_{wpa}
&=[\max(0,\rho_l-r_{bg})]^2
 +[\max(0,r_{bg}-\rho_u)]^2,\\
\mathcal{L}
&=\mathcal{L}_{act}+\lambda_{wpa}\mathcal{L}_{wpa}.
\end{aligned}
\end{equation}
In the loss, $\theta_{bg}^{(0)}$ is the initialization of the background encoder, $\theta_{bg}$ is the current encoder parameter vector, and $r_{bg}$ is their normalized distance. $\rho_l$ and $\rho_u$ define the desired adaptation interval: if the encoder changes too little, WPM may stay too generic; if it changes too much, the cached memory may become unstable. We set $\epsilon=10^{-8}$, $\rho_l=0.05$, $\rho_u=0.20$, and $\lambda_{wpa}=0.01$.

\subsection{Choice of Haar Basis}

Haar is used as the default basis because it has the shortest temporal support and therefore localizes abrupt contact changes with minimal boundary padding. This choice does not assume that real manipulation contains only constant positions or constant velocities. Acceleration, jerk, and contact-speed changes appear as localized detail coefficients, especially in $\mathbf{V}^{(1)}$ and $\mathbf{V}^{(2)}$. Longer wavelets have stronger polynomial approximation properties, but their wider support mixes a larger temporal neighborhood and can smear short contact events across several coefficients. We therefore treat Haar as a practical locality-efficiency trade-off rather than as a complete physical motion model, and we report a wavelet-family ablation in Table~\ref{tab:wavelet_basis}.
\end{revision}

\begin{revision}
\begin{table*}[t]
\vspace{5mm}
\centering
\caption{\zj{Comparison of Wavelet Policy with five baseline models. Success rate (\%) $\uparrow$. Each method is evaluated over 10 random seeds, and results are reported as mean $\pm$ standard deviation.}}
\label{tab:wavelet_main}
\resizebox{\textwidth}{!}{%
\setlength{\tabcolsep}{7pt}
\renewcommand{\arraystretch}{1.15}

\begin{tabular}{@{}lccc ccc ccc ccc@{}}
\toprule
\multicolumn{1}{c}{} &
\multicolumn{3}{c}{Transfer Cube (Sim)} &
\multicolumn{3}{c}{Bimanual Insertion (Sim)} &
\multicolumn{3}{c}{Transfer Plus (Sim)} &
\multicolumn{3}{c}{Stack Two Blocks (Sim)} \\

\cmidrule(lr){2-4}\cmidrule(lr){5-7}\cmidrule(lr){8-10}\cmidrule(l){11-13}
 & Touch & Lift & Transfer & Grasp & Contact & Insert &
   Lift & Stack & Finish & Stack & Lift & Finish \\ \midrule


DP (DDPM, CNN) ~\cite{chi2023diffusion}
 & \valpm{95.1}{1.2} & \valpm{92.4}{1.3} & \valpm{90.2}{1.9}
 & \valpm{77.4}{2.7} & \valpm{69.0}{3.1} & \valpm{63.2}{3.3}
 & \valpm{62.1}{3.5} & \valpm{52.9}{3.9} & \valpm{52.9}{3.9}
 & \valpm{82.0}{2.7} & \valpm{63.5}{3.2} & \valpm{46.7}{3.6} \\

ACT~\cite{zhao2023learning}
 & \valpm{98.4}{0.9} & \valpm{96.2}{1.0} & \valpm{94.8}{2.1}
 & \valpm{81.3}{2.6} & \valpm{73.5}{3.0} & \valpm{68.1}{3.2}
 & \valpm{66.7}{3.6} & \valpm{57.5}{4.0} & \valpm{57.5}{4.0}
 & \valpm{85.8}{2.8} & \valpm{67.3}{3.4} & \valpm{50.4}{3.8} \\[2pt]

HACT-VQ~\cite{park2024hierarchical}
 & \valpm{98.5}{0.9} & \valpm{97.6}{1.2} & \valpm{96.2}{1.4}
 & \valpm{87.4}{2.4} & \valpm{82.2}{2.7} & \valpm{76.3}{2.8}
 & \valpm{79.2}{2.2}
 & \valpm{68.5}{3.1} & \valpm{68.5}{3.1}
 & \valpm{90.6}{2.3} & \valpm{76.1}{2.8} & \valpm{55.8}{3.2} \\[2pt]

InterACT~\cite{lee2025interact}
 & \valpm{98.2}{0.8}
 & \valpm{88.4}{1.1} & \valpm{82.1}{2.0}
 & \valpm{88.5}{2.2}
 & \valpm{78.3}{2.8} & \valpm{44.2}{3.2}
 & \valpm{78.5}{2.4} & \valpm{68.7}{3.4} & \valpm{68.7}{3.4} 
 & \valpm{91.9}{2.0} & \valpm{77.1}{2.6} & \valpm{56.9}{2.9} \\[2pt]
 
GR00T N 1.7~\cite{gr00tn1_2025}
 & \valpm{94.5}{1.5} & \valpm{91.2}{2.0} & \valpm{91.6}{2.2}
 & \valpm{83.2}{2.5} & \valpm{73.9}{3.0} & \valpm{69.1}{3.3}
 & \valpm{63.4}{3.1} & \valpm{57.4}{3.9} & \valpm{57.4}{3.9}
 & \valpm{81.3}{2.9} & \valpm{65.4}{3.2} & \valpm{47.5}{3.1} \\[2pt]

\cellcolor[HTML]{CCF2F5}Ours
 & \cellcolor[HTML]{CCF2F5}\textbf{\valpm{99.5}{0.3}}
 & \cellcolor[HTML]{CCF2F5}\textbf{\valpm{98.3}{0.4}}
 & \cellcolor[HTML]{CCF2F5}\textbf{\valpm{97.1}{0.5}}
 & \cellcolor[HTML]{CCF2F5}\textbf{\valpm{89.5}{1.5}}
 & \cellcolor[HTML]{CCF2F5}\textbf{\valpm{88.4}{1.8}}
 & \cellcolor[HTML]{CCF2F5}\textbf{\valpm{82.2}{2.0}}
 & \cellcolor[HTML]{CCF2F5}\textbf{\valpm{81.8}{2.2}}
 & \cellcolor[HTML]{CCF2F5}\textbf{\valpm{72.4}{2.1}}
 & \cellcolor[HTML]{CCF2F5}\textbf{\valpm{72.4}{2.1}}
 & \cellcolor[HTML]{CCF2F5}\textbf{\valpm{92.4}{1.8}}
 & \cellcolor[HTML]{CCF2F5}\textbf{\valpm{74.3}{2.0}}
 & \cellcolor[HTML]{CCF2F5}\textbf{\valpm{68.4}{2.4}} \\
\bottomrule
\end{tabular}
}
\vspace{-2mm}
\end{table*}

\begin{table*}[t]
\centering
\caption{\zj{Real-world evaluation on physical manipulation tasks.} Success rates are reported as mean $\pm$ 95\% confidence interval over 150 rollouts per task stage.}
\label{tab:wavelet_reale}
{
\resizebox{0.96\textwidth}{!}{%
\begin{tabular}{@{}lccc ccc ccc@{}}
\toprule
& \multicolumn{3}{c}{Stack Block} & \multicolumn{3}{c}{Store Strawberry} & \multicolumn{3}{c}{Store Lemon} \\
\cmidrule(lr){2-4}\cmidrule(lr){5-7}\cmidrule(l){8-10}
Method & Grasp & Lift & Stack & Grasp & Lift & Place & Grasp & Lift & Place \\ \midrule
ACT~\cite{zhao2023learning}
 & \valpm{0.85}{0.06} & \valpm{0.74}{0.07} & \valpm{0.69}{0.07}
 & \valpm{0.93}{0.04} & \valpm{0.85}{0.06} & \valpm{0.74}{0.07}
 & \valpm{0.76}{0.07} & \valpm{0.67}{0.08} & \valpm{0.65}{0.08} \\
InterACT~\cite{lee2025interact}
 & \valpm{0.95}{0.03} & \valpm{0.81}{0.06} & \valpm{0.68}{0.07}
 & \valpm{0.88}{0.05} & \valpm{0.82}{0.06} & \valpm{0.79}{0.07}
 & \valpm{0.85}{0.06} & \valpm{0.75}{0.07} & \valpm{0.67}{0.08} \\
\rowcolor[HTML]{CCF2F5}\textbf{Ours}
 & \textbf{\valpm{0.92}{0.04}} & \textbf{\valpm{0.88}{0.05}} & \textbf{\valpm{0.80}{0.06}}
 & \textbf{\valpm{0.97}{0.03}} & \textbf{\valpm{0.93}{0.04}} & \textbf{\valpm{0.88}{0.05}}
 & \textbf{\valpm{0.92}{0.04}} & \textbf{\valpm{0.89}{0.05}} & \textbf{\valpm{0.78}{0.07}} \\
\bottomrule
\end{tabular}}}

\vspace{1mm}
{
\resizebox{0.96\textwidth}{!}{%
\begin{tabular}{@{}lccc ccc ccc@{}}
\toprule
& \multicolumn{3}{c}{Store Items} & \multicolumn{3}{c}{Assist Sewing} & \multicolumn{3}{c}{Stack Blocks} \\
\cmidrule(lr){2-4}\cmidrule(lr){5-7}\cmidrule(l){8-10}
Method & First & Second & Finish & Contact & Feed & Align & Stack & Lift & Finish \\ \midrule
ACT~\cite{zhao2023learning}
 & \valpm{0.86}{0.06} & \valpm{0.65}{0.08} & \valpm{0.58}{0.08}
 & \valpm{0.87}{0.05} & \valpm{0.81}{0.06} & \valpm{0.68}{0.07}
 & \valpm{0.65}{0.08} & \valpm{0.56}{0.08} & \valpm{0.47}{0.08} \\
InterACT~\cite{lee2025interact}
 & \valpm{0.77}{0.07} & \valpm{0.74}{0.07} & \valpm{0.68}{0.07}
 & \valpm{0.86}{0.06} & \valpm{0.74}{0.07} & \valpm{0.67}{0.08}
 & \valpm{0.78}{0.07} & \valpm{0.67}{0.08} & \valpm{0.61}{0.08} \\
\rowcolor[HTML]{CCF2F5}\textbf{Ours}
 & \textbf{\valpm{0.94}{0.04}} & \textbf{\valpm{0.88}{0.05}} & \textbf{\valpm{0.76}{0.07}}
 & \textbf{\valpm{0.95}{0.03}} & \textbf{\valpm{0.85}{0.06}} & \textbf{\valpm{0.74}{0.07}}
 & \textbf{\valpm{0.87}{0.05}} & \textbf{\valpm{0.75}{0.07}} & \textbf{\valpm{0.70}{0.07}} \\
\bottomrule
\end{tabular}}}
\vspace{-3mm}
\end{table*}
\end{revision}


\section{Experiments}
\label{sec:experiments}
\begin{revision}


We evaluate Wavelet Policy by asking three complementary questions.
\textbf{Q1: Does Wavelet Policy outperform strong visuomotor imitation learning baselines?}
To answer this question, we compare Wavelet Policy with representative baselines on both simulated and real-world manipulation tasks.
\textbf{Q2: Does wavelet-domain action modeling help fine-grained manipulation?}
To answer this question, we study the adaptation behavior and fine-tuning efficiency when transferring from a simpler cube-transfer task to a more precision-demanding stacking task.
\textbf{Q3: Does WPM improve robustness under distribution shift?}
To answer this question, we evaluate the role of WPM under appearance and occlusion shifts, where the test-time visual distribution differs from the training distribution.

\begin{figure*}[t]
    \centering
    \includegraphics[width=0.92\textwidth]{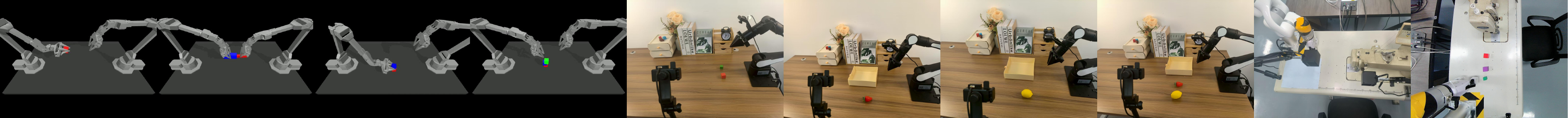}
    \caption{
        Left: The MuJoCo tasks \textit{Transfer Cube}, \textit{Bimanual Insertion}, \textit{Transfer Plus}, and \textit{Stack Two Blocks}. Right: The real-world tasks \textit{Stack Block}, \textit{Store Strawberry}, \textit{Store Lemon}, \textit{Store Items}, \textit{Assist Sewing}, and \textit{Stack Blocks}.}
    \label{fig:eperiment}
\end{figure*}

\subsection{Experimental Setup}
We evaluated Wavelet Policy on four simulated robotic arm tasks and six real-world tasks, as illustrated in Fig.~\ref{fig:eperiment}. In the simulation benchmarks, Transfer Cube and Bimanual Insertion are taken from~\cite{zhao2023learning}, while Transfer Plus and Stack Two Blocks are new, more challenging tasks we designed. Transfer Plus extends cube transfer by requiring the robot to stack the transferred cube onto another block, and Stack Two Blocks requires two arms to sequentially stack two blocks together.

All policies predict $H=40$ future actions at a 20 Hz control frequency. In bimanual simulation, the action dimension is $D=14$, corresponding to the joint commands and gripper commands of two 6-DoF arms. In the real-world single-arm tasks, $D=7$, corresponding to 6 arm joints and one gripper command. Simulation objects include rigid cubes or blocks with side lengths of approximately 3--5 cm and insertion fixtures with 1--2 cm clearance. The tabletop workspace is approximately $0.60\,\mathrm{m}\times0.45\,\mathrm{m}$, with object initial positions sampled within a 10--25 cm reachable range from the nominal grasp pose. Unless a task has a stricter semantic condition, a rollout is considered successful when the final object position is within 2 cm of the target region and the relevant orientation error is below $10^\circ$.

The real-world task suite spans two robotic platforms: the Airbot Play academic arm from DISCOVER Robotics and the UFactory xArm6 industrial manipulator. RGB observations are captured at $640\times480$ and resized to $480\times640$ for the policy. The real workspace is approximately $0.55\,\mathrm{m}\times0.40\,\mathrm{m}$ for tabletop tasks and $0.70\,\mathrm{m}\times0.45\,\mathrm{m}$ for the sewing-assist setup. We use rigid toy blocks, plastic fruit props, storage containers, and fabric-feeding fixtures with characteristic sizes of 3--12 cm. Demonstrations for Airbot Play are collected by human teleoperation. For the xArm6 industrial-arm setting, we use kinesthetic or scripted waypoint execution with vision-based alignment and manually verify each recorded trajectory before adding it to the training set.

For Wavelet Policy, we use a three-level wavelet decomposition, i.e., $S=3$, which produces one approximation branch and three detail branches $\mathcal{W}(\mathbf{Z})
=
\{\mathbf{Z}^{A_3}, \mathbf{Z}^{D_3}, \mathbf{Z}^{D_2}, \mathbf{Z}^{D_1}\},$
where $A_3$ and $D_3$ mainly describe coarse motion structure, whereas $D_2$ and $D_1$ capture middle- and high-frequency local corrections that are critical for contact-rich alignment and final stacking. Unless stated otherwise, Wavelet Policy uses a single camera view. 

For a fair comparison, all task-specific policies, including ACT, InterACT, HACT-VQ, Diffusion Policy, GR00T N1.7, and Wavelet Policy, are trained and evaluated using the same demonstration datasets, camera observations, image resolution, action horizon, control frequency, validation split, checkpoint-selection criterion, and rollout execution protocol for each task. Diffusion Policy is evaluated using DDPM with 100 denoising steps under the same action-chunk setting. For GR00T N1.7, we use the same visual observations and robot action space, provide the task instruction as a language prompt, and perform task-specific post-training on the same demonstrations before evaluation. Upon publication, we will release the complete codebase, task environments, configuration files, data-generation protocols, and evaluation scripts to support reproducibility. Except as stated above, for all baseline methods, unless otherwise specified, we follow the default configurations reported in their original paper or official implementation, including network architecture, training hyperparameters, and inference settings.

Training was performed on NVIDIA RTX 3090 and RTX 4090 GPUs. Inference timing was measured on an NVIDIA GTX 1650 and an RTX 4090. These measurements characterize lightweight GPU inference under resource-constrained settings; evaluation on embedded boards such as Jetson devices remains future work.

\begin{table}[t]
\centering
\caption{\zj{Key hyperparameters used for Wavelet Policy unless otherwise specified.}}
\label{tab:hyperparams}
\resizebox{\columnwidth}{!}{%
\begin{tabular}{@{}lc@{}}
\toprule
Item & Value \\ \midrule
Input resolution & $480\times640$ RGB \\
Action horizon $H$ & 40 \\
Wavelet levels $S$ & 3 \\
Visual token length $L$ & 39 per camera \\
Embedding width after FE & $64N_c$ before projection \\
Optimizer & AdamW \\
Learning rate & $1\times10^{-4}$ \\
Weight decay & $1\times10^{-4}$ \\
Batch size & 32 (simulation), 16 (real world) \\
Training iterations & 100k \\
Loss & L1 action loss + $\lambda_{wpa}\mathcal{L}_{wpa}$ \\
$\lambda_{wpa}$ & 0.01 \\
Transformer dropout & 0.1 \\
\bottomrule
\end{tabular}}
\vspace{-3mm}
\end{table}

\subsection{Main Results}
\subsubsection{Simulation Evaluation}
\label{sec:main_results}
All simulation, fine-tuning, and WPM robustness experiments average 10 random seeds. Each seed is trained for 100k iterations on 100 episodes and evaluated on 100 test episodes using the best validation checkpoint.

We compare with Diffusion Policy (DDPM, CNN)~\cite{chi2023diffusion}, ACT~\cite{zhao2023learning}, InterACT~\cite{lee2025interact}, HACT-VQ~\cite{park2024hierarchical}, and GR00T N 1.7~\cite{gr00tn1_2025}. As shown in Table~\ref{tab:wavelet_main}, Wavelet Policy achieves the best final success on all four simulation tasks: $97.1\%$ on \textit{Transfer Cube}, $82.2\%$ on \textit{Bimanual Insertion}, $72.4\%$ on \textit{Transfer Plus}, and $68.4\%$ on \textit{Stack Two Blocks}. The gains are largest in tasks requiring both long-range transport and contact-sensitive refinement. Table~\ref{table:se2md} further shows lower inference latency than ACT under one- and two-camera inputs.

\begin{table}[t]
\vspace{2mm}
\centering
\caption{\zj{Per-frame inference average latency under single- and dual-camera inputs on GTX 1650. Latency (s/frame) $\downarrow$.}}
\resizebox{0.49\textwidth}{!}{%
\begin{tabular}{@{}cccc@{}}
\toprule
\multirow{1}{*}{Model} 
                       & Params           & One Cam & Two Cam \\ \midrule
DP (DDPM, CNN)~\cite{chi2023diffusion} & 308.29 M          & 0.1025               & 0.1201               \\
DP (DDIM)~\cite{fu2024mobile}          & 77.55 M           & 0.0604               & 0.0785               \\
ACT                    & 83.92 M           & 0.0390               & 0.0480               \\
InterACT~\cite{lee2025interact}       & 67.70 M           & 0.0380               & 0.0423               \\
HACT-VQ~\cite{park2024hierarchical}   & 125.49 M          & 0.0685               & 0.0842               \\
GR00T N 1.7~\cite{gr00tn1_2025}       & 3.00 B           & 0.2204               & 0.2496               \\
$\pi_{0.7}$~\cite{intelligence2026pi}   & 23.66 B          & -               & -               \\
Ours (one cam)         & \textbf{18.20 M}  & 0.0308               & -               \\
Ours (two cams)        & 25.31 M     & -               & 0.0331               \\ \bottomrule

\end{tabular}}
\label{table:se2md}
\end{table}

\begin{table}[t]
\centering
\caption{\zj{Wavelet-basis ablation on representative contact-rich simulation tasks.} Success is reported as mean $\pm$ standard deviation over 10 seeds.}
\label{tab:wavelet_basis}
\resizebox{\columnwidth}{!}{%
\begin{tabular}{@{}lcc@{}}
\toprule
Basis & Bimanual Insert & Transfer Plus Finish \\ \midrule
Haar (db1) & \textbf{\valpm{78.3}{2.8}} & \textbf{\valpm{70.4}{3.1}} \\
bior1.1 & \valpm{75.0}{3.1} & \valpm{67.0}{3.3} \\
bior2.4 & \valpm{76.1}{3.0} & \valpm{68.2}{3.1} \\
rbio1.1 & \valpm{73.9}{3.3} & \valpm{65.8}{3.3} \\
sym5 & \valpm{75.7}{3.0} & \valpm{67.6}{3.1} \\
db5 & unstable & unstable \\
coif3 & unstable & unstable \\
\bottomrule
\end{tabular}}
\vspace{-3mm}
\end{table}

\begin{figure*}[t]
\vspace{5mm}
\centering
\includegraphics[width=0.92\textwidth]{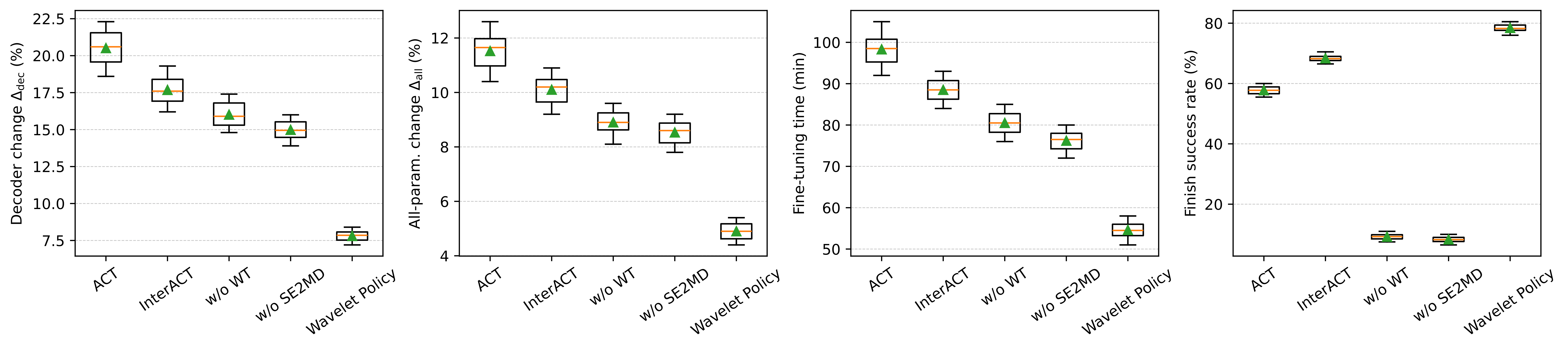}
\caption{
Overall fine-tuning analysis from \textit{Transfer Cube (Sim)} to \textit{Transfer Plus (Sim)}.
}
\label{fig:finetune_transfer_overall_boxplot}
\end{figure*}

\begin{figure*}[t]
    \centering
    \includegraphics[width=0.90\textwidth]{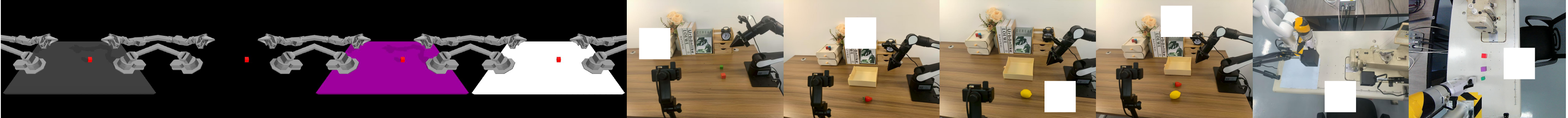}
    \caption{
        Examples of WPM inputs under distribution-shift settings. 
        The left four panels show simulation backgrounds with default, black, purple, and white table appearances, respectively. 
        The right six panels show representative real-world initial frames from six manipulation tasks, where a random square mask covering $5\%$ of the image area is applied to the WPM input.
}
    \label{fig:eperiment_plus}
\end{figure*}

\subsubsection{Real-World Evaluation}
\label{sec:real_world_eval}
Real-world tasks are evaluated on Airbot Play and UFactory xArm6. Initial object poses are randomized within the reachable workspace, and failures include missed grasps, object drops, wrong placement, or final misalignment. Table~\ref{tab:wavelet_reale} reports three seeds with 50 physical rollouts per seed, giving 150 rollouts per task stage and 95\% binomial confidence intervals. Wavelet Policy consistently outperforms ACT and InterACT, especially in final placement and alignment stages. Diffusion Policy is kept in simulation and latency comparisons because 100-step DDPM inference does not meet our 20 Hz physical control period.

\subsection{Does Wavelet Modeling Help Fine-Grained Manipulation?}
\label{sec:fine_grained_wavelet}

To examine whether wavelet-domain modeling benefits fine-grained manipulation, we fine-tune policies from \textit{Transfer Cube (Sim)} to \textit{Transfer Plus (Sim)}, which adds a precision-demanding stacking stage after the shared cube-transfer behavior. We compare ACT, InterACT, Wavelet Policy without WT, Wavelet Policy without SE2MD, and the full model. Adaptation is measured by decoder parameter change, whole-model parameter change, wall-clock fine-tuning time, final success rate, and the fine-decoder update share $R_{\mathrm{fine}}$, defined as the fraction of scale-decoder parameter update norm assigned to $D_1$ and $D_2$. A larger $R_{\mathrm{fine}}$ indicates stronger adaptation in the local-refinement branches.

\begin{figure}[t]
\vspace{5mm}
\centering
\includegraphics[width=\columnwidth]{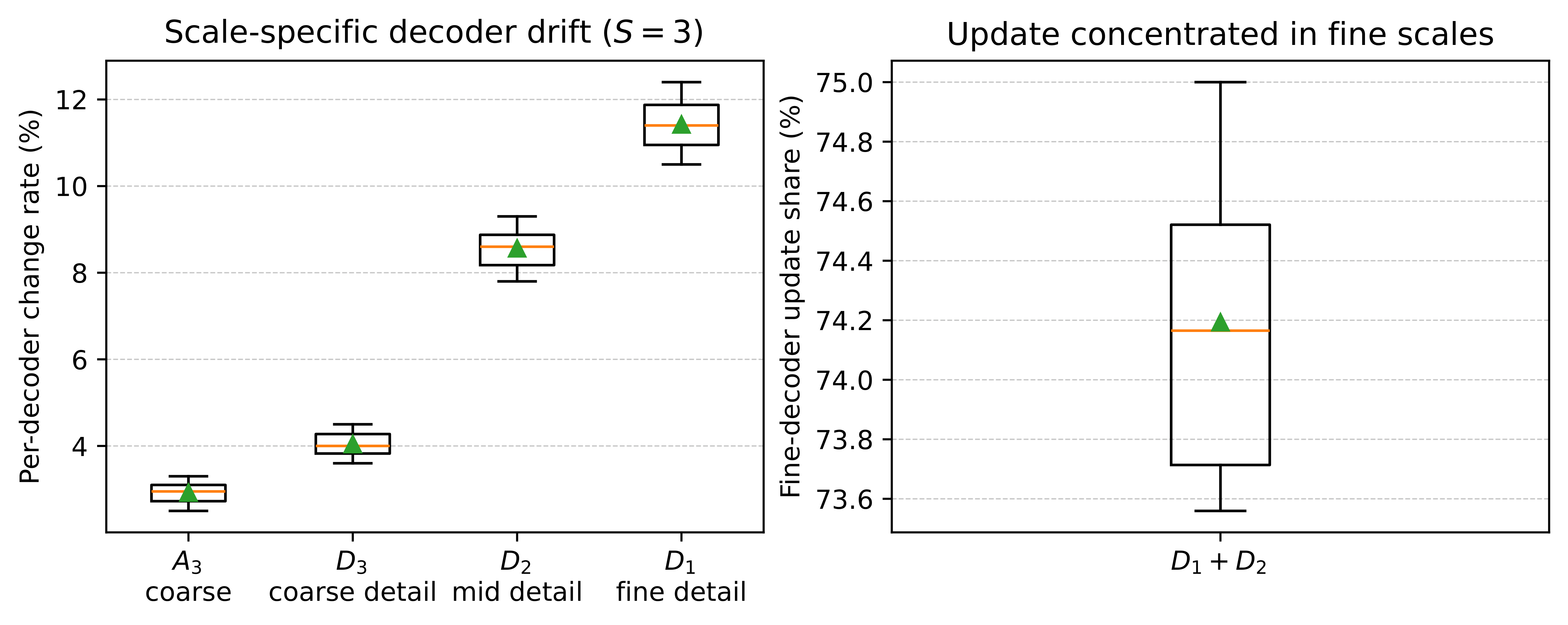}
\caption{
Scale-specific decoder adaptation of Wavelet Policy with $S=3$.
}
\label{fig:finetune_scale_decoder_boxplot}
\end{figure}

\begin{figure*}[t]
    \centering
    \includegraphics[width=0.92\textwidth]{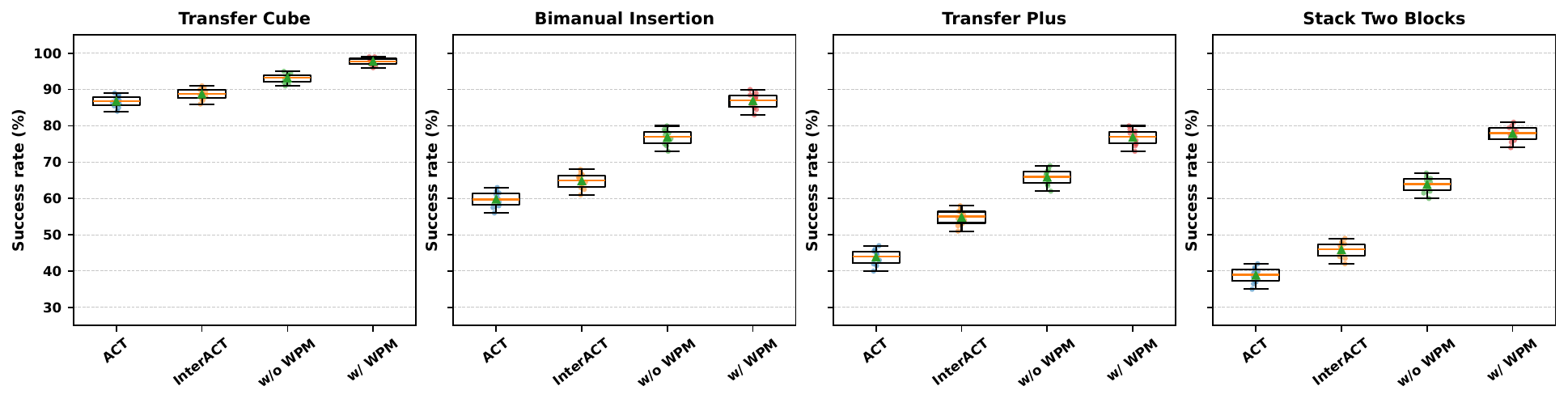}
    \caption{
    WPM evaluation under table-color shift in simulation.
    }
    \label{fig:wpm_color_shift}
\end{figure*}
\begin{figure*}[t]
\vspace{5mm}
    \centering
    \includegraphics[width=0.92\textwidth]{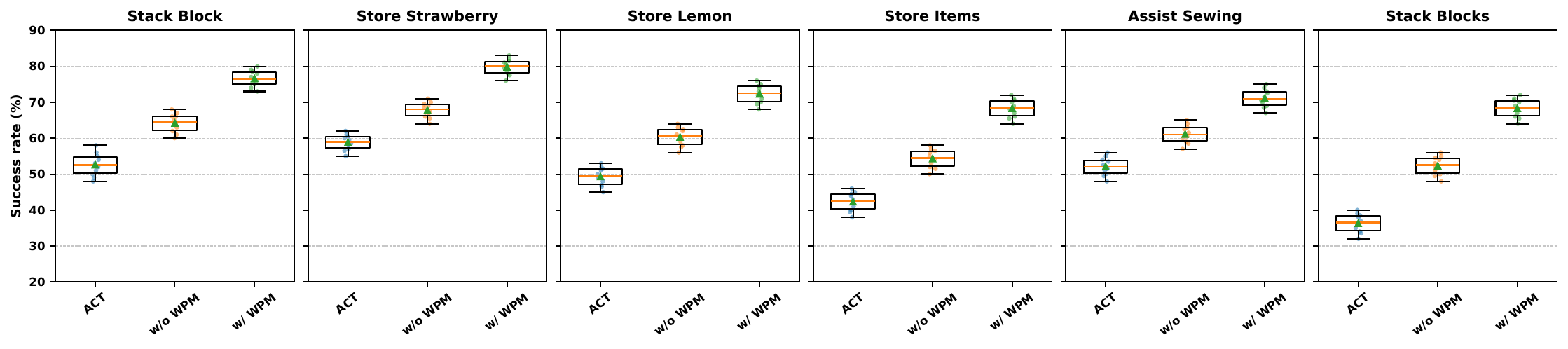}
    \caption{
    WPM evaluation under background-occlusion shift in real-world tasks.
    }
    \label{fig:wpm_background_occlusion}
\end{figure*}
As shown in Fig.~\ref{fig:finetune_transfer_overall_boxplot}, Wavelet Policy needs smaller parameter changes and less fine-tuning time while achieving higher final success. Fig.~\ref{fig:finetune_scale_decoder_boxplot} further shows that updates concentrate in $D_1$ and $D_2$, while coarse branches $A_3$ and $D_3$ change less. This supports the view that the new stacking stage mainly requires local contact refinement, and that both WT and SE2MD help isolate this adaptation.

\subsection{Does WPM Improve Robustness under Distribution Shift?}
\label{sec:wpm_generalization}

We further isolate WPM under visual distribution shifts. Before each rollout, the initial image is encoded once and cached as the WPM input; in \textit{Transfer Cube (Sim)}, this image is captured before the cube is placed in the scene.

\subsubsection{Table-Color Shift in Simulation}

In simulation, table color is sampled from black and purple during training and changed to unseen white during testing. The WPM input is the first frame of each episode, so it captures static workspace layout and appearance without leaking the manipulated cube position.

Fig.~\ref{fig:wpm_color_shift} shows that WPM improves robustness under the unseen white-table condition by providing persistent scene-level cues such as table geometry and workspace boundaries.

\subsubsection{Background-Occlusion Shift in Real-World Tasks}

We also corrupt only the WPM background image in real-world tasks, leaving online observations unchanged. During training, a random square mask covers $5\%$ of the WPM image; during testing, the mask increases to $6\%$--$10\%$ with random location.

As shown in Fig.~\ref{fig:wpm_background_occlusion}, WPM maintains stronger performance under partial background occlusion, which is important when lighting changes, camera noise, or occlusions corrupt static scene observations.

These two stress tests suggest that WPM is not simply memorizing object locations. In both settings, the manipulated object is either absent from the WPM frame or the perturbation is applied only to the cached background input. The improvement is therefore more consistent with a static-scene prior: the policy receives a compact reminder of table boundaries, support surfaces, fixtures, and non-interactive regions, while the online observation remains responsible for the current object state. This separation is useful for lightweight policies because it adds persistent scene context without increasing per-step visual encoding cost.
\end{revision}

\section{CONCLUSION AND LIMITATION}

\begin{revision}
In this work, we present Wavelet Policy, a lightweight visuomotor imitation learning framework that combines wavelet-domain action modeling with WPM. By decomposing latent future-action tokens into multi-scale components, our method separates coarse motion trends from fine-grained local corrections. WPM further injects persistent scene-level priors from background images, enhancing robustness under appearance and background shifts while adding little online inference cost. Experiments show that Wavelet Policy consistently outperforms strong baselines with better accuracy and efficiency.

This work also has limitations. WPM assumes that the static background provides useful and relatively stable scene information, which may be weakened in highly dynamic environments or tasks with moving fixtures. In addition, our current wavelet design uses a fixed decomposition depth and a fixed Haar basis; although this choice is efficient and empirically stable, adaptive basis or scale selection remains unexplored. Finally, the current latency results characterize lightweight deployment on GPU hardware, while more complete embedded-board evaluation is left for future work. Future work will extend Wavelet Policy to more dynamic scenes and investigate task-adaptive wavelet representations.
\end{revision}

\bibliographystyle{IEEEtran}
\bibliography{IEEEexample}
\end{document}